\documentclass[lettersize,journal]{IEEEtran}
\usepackage{amsmath,amsfonts,bm}
\usepackage{amssymb}
\usepackage{algorithmic}
\usepackage{algorithm}
\usepackage{array}
\usepackage[caption=false,font=normalsize,labelfont=sf,textfont=sf]{subfig}
\usepackage{textcomp}
\usepackage{stfloats}
\usepackage{url}
\usepackage{verbatim}
\usepackage{graphicx}
\usepackage{cite}
\usepackage{booktabs}
\usepackage{multirow}  
\usepackage{graphicx}  
\usepackage{makecell}
\usepackage{diagbox}
\usepackage{epstopdf}
\usepackage[caption=false]{subfig}
\usepackage{caption}
\newcommand{\Rmnum}[1]{\uppercase\expandafter{\romannumeral #1}}  
\begin{document}

\title{Accurate Network Traffic Matrix Prediction via \textit{LEAD}: a Large Language Model-Enhanced Adapter-Based Conditional Diffusion Model}

\author{Yu Sun,
	Yaqiong Liu,~\IEEEmembership{Senior Member,~IEEE,}
	Nan Cheng,\\
	Jiayuan Li,
	Zihan Jia,
	Xialin Du,
	and~Mugen Peng,~\IEEEmembership{Fellow,~IEEE}
	\thanks{Y. Sun, Y. Liu (Corresponding author), X. Du and M. Peng are with the School of Information and Communication Engineering, Beijing University of Posts and Telecommunications, Beijing 100876, China (e-mail: 2024110229@bupt.cn; liuyaqiong@bupt.edu.cn; XialinDu@bupt.edu.cn; pmg@bupt.edu.cn)}
	\thanks{N. Cheng, J. Li, and Z. Jia are with the China Mobile Group Design Institute, Beijing 100080, China (e-mail: chengnan@cmdi.chinamobile.com; lijiayuan@cmdi.chinamobile.com; jiazihan@cmdi.chinamobile.com)}
	\thanks{Manuscript received April 19, 2025; revised August 16, 2025.}}

\markboth{Journal of \LaTeX\ Class Files,~Vol.~14, No.~8, August~2026}%
{Sun \MakeLowercase{\textit{et al.}}:Accurate Network Traffic Matrix Prediction via \textit{LEAD}: a Large Language Model-Enhanced Adapter-Based Conditional Diffusion Model}


\maketitle

\begin{abstract}
In the era of AI-native 6G, network operations are evolving from reactive, rule-based paradigms to proactive, cognitive frameworks. To realize this vision, accurate Network Traffic Matrix (TM) prediction serves as a critical sensing mechanism, providing the situational awareness necessary for intelligent resource orchestration. However, accurate TM forecasting remains challenging due to the stochastic, non-linear, and bursty nature of network dynamics. Existing discriminative models often suffer from over-smoothing and provide limited uncertainty awareness, leading to poor fidelity under extreme bursts. To address these limitations, we propose \textbf{\textit{\textit{LEAD}}}, a Large Language Model-Enhanced Adapter-based conditional Diffusion model. First, \textit{LEAD} adopts a \textbf{``Traffic-to-Image''} paradigm to transform traffic matrices into RGB images, enabling global dependency modeling via vision backbones. Then, we design a ``\textbf{Frozen LLM with Trainable Adapter}'' model, which efficiently captures temporal semantics with limited computational cost. Moreover, we propose a \textbf{Dual-Conditioning Strategy} to precisely guide a diffusion model to generate complex, dynamic network traffic matrices. Experiments on the Abilene and GÉANT datasets demonstrate that \textit{LEAD} outperforms all baselines. On the Abilene dataset, \textit{LEAD} attains a remarkable 45.2\% reduction in RMSE against the best baseline, with the error margin rising only marginally from 0.1098 at one-step to 0.1134 at 20-step predictions. Meanwhile, on the GÉANT dataset, \textit{LEAD} achieves a 0.0258 RMSE at 20-step prediction horizon which is 27.3\% lower than the best baseline. 
\end{abstract}


\section{Introduction}\label{sec:introduction}
\IEEEPARstart{A}{s} wireless systems evolve toward 6G communication, network operations increasingly require cognitive, closed-loop predictive adaptation under stringent latency and resource constraints across distributed edge nodes. Meanwhile, the explosive growth of Internet traffic, driven by emerging applications like ultra-high-definition video streaming, cloud computing, and the Internet of Things (IoT), continues to place unprecedented demands on modern communication networks \cite{Liu25_Survey_GenAI_6G}, \cite{Qin25_GenAI_IntentDriven}, \cite{Zhang25_GC_FL_Traffic}. Central to effective network management is the network Traffic Matrix (TM), which captures the time-varying aggregate flow volume exchanged between network nodes over a given time interval. By offering a global view of network demand, TMs serve as the cornerstone for critical tasks such as Traffic Engineering (TE), Capacity Planning, and Service Level Agreement (SLA) assurance \cite{Vardi96}, \cite{Sun25_TME_LinkLoadSampling}, \cite{Lv24_KET_FL}. In these operational loops, accurately anticipating bursts and traffic volatility is critical for risk-aware actions rather than purely reactive responses. However, obtaining real-time TMs is often plagued by high measurement overhead and significant transmission delays, making reactive adjustments insufficient for guaranteeing Quality of Service (QoS) in modern high-speed networks. Consequently, TM prediction, defined as forecasting network traffic loads based on historical measurements, has emerged as an indispensable capability because it empowers network operators to proactively optimize routing protocols and mitigate congestion before it physically occurs, shifting network management from a reactive to a proactive paradigm.

Traditional TM prediction often uses statistical time-series and regression models such as ARIMA (Autoregressive Integrated Moving Average) and Kalman filtering \cite{Zhang24_AIKF_TMMonitoring}. While efficient, they typically rely on simplified assumptions and weakly capture network-wide spatial coupling and multi-timescale temporal dynamics, which can reduce robustness under bursty traffic in large backbone networks.

Deep learning methods provide a natural remedy by learning richer temporal dependencies from historical measurements. RNNs (Recurrent Neural Networks) improve sequential modeling but may still miss spatial interactions and long-range effects. Methods based on GNN (Graph Neural Network) address this by encoding adjacency structure, and recent work further explores spatio-temporal attention for TM modeling \cite{Liu25_SAMS_GNN}. However, they can be sensitive to graph specification and over-smoothing, weakening responsiveness to peaks and anomalies. Moreover, most existing predictors are deterministic and provide only point forecasts, thus failing to quantify uncertainty and limiting their utility for risk-aware operation under strict SLAs.

A promising shift is to represent traffic matrices as images \cite{Kablaoui24}. By mapping TM entries to pixel intensities, spatial patterns such as network hotspots and directional flows can be more effectively captured. This representation allows researchers to leverage powerful, pretrained vision models, which have learned robust hierarchical features from large-scale datasets like ImageNet \cite{He16}. However, directly applying standard vision encoders introduces a domain mismatch in that network traffic is sparse and non-smooth, and its dependencies are governed by topology and routing rather than visual locality. Consequently, vision features alone may not faithfully reflect network-specific semantics such as temporal causality, diurnal regularities, or protocol-driven bursts.

To move beyond deterministic regression, we consider conditional diffusion models, which support distribution modeling and uncertainty quantification \cite{Ho20}. Although diffusion has shown promise in TM-related network-management tasks \cite{10835411}, accurate forward prediction remains challenging when conditioning is shallow and fails to express higher-level temporal semantics in highly stochastic, bursty traffic.

We therefore hypothesize that Large Language Models (LLMs) can provide semantic abstraction and reasoning to strengthen conditioning for generative TM forecasting. However, directly fine-tuning massive LLMs for a niche network-traffic task is often computationally prohibitive, so we adopt parameter-efficient adaptation inspired by the LLaMA-Adapter \cite{Zhang24}, which enables the efficient alignment of frozen LLMs with the network traffic domain via lightweight adaptation.

In this paper, we propose a novel model, \textbf{{\textit{LEAD}} (\underline{L}LM-\underline{E}nhanced \underline{A}dapter-based conditional \underline{D}iffusion model)}, which synergistically combines the semantic reasoning of a frozen LLM with the generative power of a conditional diffusion model. Our approach processes historical TMs through a tailored pipeline where a frozen LLM acts as a semantic encoder to distill underlying traffic evolution patterns, such as daily periodicities and unexpected bursts, which then guide a diffusion U-Net to generate high-fidelity future traffic matrices. 

Our contributions are summarized as follows:

\begin{itemize}
	\item We propose \textbf{\textit{\textit{LEAD}}}, a novel model that orchestrates a deep synergy between LLMs and conditional diffusion models for network traffic prediction. Beyond simple integration, we establish a dependency where the LLM acts as a semantic reasoner to extract high-level temporal dynamics. This logic serves as explicit guidance for the diffusion model, ensuring the generated traffic matrices are not only numerically precise but also consistent with realistic network behaviors and topological constraints, while keeping the LLM deployable via parameter-efficient adaptation for proactive and self-adaptive cognitive management.
	
	\item We represent the network traffic matrices with RGB images and design a specialized vision encoder that is specifically tailored to handle the sparsity and high-frequency characteristics of network traffic data.
	
	\item We design a dual-scale conditioning strategy (global and sequential conditions) within the diffusion process, ensuring the generated traffic matrices maintain both macroscopic structural consistency and microscopic temporal fidelity.
	
	\item We conduct comprehensive experiments on real-world datasets Abilene and GÉANT, and our model demonstrates substantial improvements in prediction accuracy.
\end{itemize}

The remainder of this paper is organized as follows: Section \ref{sec:relatedworks} reviews related works. Section \ref{sec:methodology} details the methodology of the proposed \textit{\textit{LEAD}} model. Section \ref{sec:experiment} reports our experimental results, including ablation studies. Section \ref{sec:conclusion} concludes the paper and proposes future research directions.

\section{Related Works}{\label{sec:relatedworks}}
\subsection{Network Traffic Prediction Methods}
The evolution of network traffic prediction methods generally progresses from statistical approaches to deep learning frameworks that integrate temporal and spatial modeling.

Early methodologies rely heavily on statistical techniques. Historical Average, ARIMA \cite{Groschwitz94}, and Kalman filtering \cite{Zhang24_AIKF_TMMonitoring} are classic examples that model linear dependencies, while Vector Auto-Regression (VAR) \cite{Duker24} extends these to multivariate settings. Despite their efficiency, these models are limited by stationarity assumptions and often struggle to capture the non-linear dynamics of complex networks.

To model non-linear temporal correlations, RNNs, such as LSTM (Long Short-Term Memory) and GRU (Gated Recurrent Unit) \cite{11185220}, are introduced and become foundational baselines. However, these methods treat network traffic streams as isolated sequences, largely neglecting the underlying topological structure and spatial interactions between nodes.

Consequently, Spatio-Temporal Graph Neural Networks (STGNNs) have been widely adopted to jointly model spatial and temporal dependencies. Representative works like DCRNN (Diffusion Convolutional Recurrent Neural Network) \cite{li2018diffusion} and STGCN (Spatial-Temporal Graph Convolutional Network) \cite{Yu18} combine graph convolutions with recurrent or convolutional units to capture directed spatial flows. To mitigate the reliance on predefined graph structures, methods such as Graph WaveNet \cite{VanDenOord16} and AGCRN (Adaptive Graph Convolutional Recurrent Network) \cite{Bai20} propose adaptive graph learning strategies to infer latent correlations automatically. Furthermore, recent studies in the database community have actively explored dynamic and multiscale structure learning to capture evolving dependencies. For instance, METRO \cite{Cui22} introduces a generic framework leveraging multiscale temporal graphs to model variable correlations, while Zhao et al. \cite{Zhao23} propose explicitly generating discrete graph structures to adapt to changing time-series relationships. In parallel, researchers have explored advanced representation learning paradigms. For instance, ST-SSL (Spatio-Temporal Self-Supervised Learning) \cite{Ji23} leverages self-supervised learning schemes to enhance robustness against data heterogeneity, while MC-STL (Mask- and Contrast-enhanced Spatio-Temporal Learning) \cite{Zhang23} employs multi-channel architectures to disentangle complex trend and seasonality components. More recently, researchers have explored incorporating domain knowledge into STGNNs through attention mechanisms \cite{11114796}.

Despite the effectiveness of these discriminative models, they typically function as deterministic regressors. This deterministic nature limits their ability to fully characterize the multi-modal probability distributions of future traffic, particularly in highly stochastic scenarios.

\subsection{Diffusion Models}
Diffusion Probabilistic Models (DPMs) have recently emerged as a powerful class of generative models, challenging the dominance of GANs (Generative Adversarial Networks) and VAEs (Variational Auto-Encoders). The core mechanism involves two processes: a forward diffusion process that gradually adds Gaussian noise to the data until it becomes pure noise, and a reverse denoising process where a neural network is trained to reconstruct the original data structure iteratively \cite{Ho20}. This paradigm allows for high-fidelity sample generation and accurate distribution modeling. More recently, diffusion models have also been explored in the context of network management, demonstrating their effectiveness for TM analysis tasks including TM synthesis, tomography, and completion \cite{10835411}.

Building on their success in computer vision, diffusion models have been increasingly adapted for time-series forecasting and imputation. TimeGrad \cite{Rasul21} is among the first to introduce diffusion to this domain, employing an autoregressive denoising process to model the conditional distribution of univariate time series. It demonstrates that diffusion-based objectives could yield superior uncertainty estimates compared to deterministic RNNs. Subsequently, CSDI (Conditional Score-based Diffusion models for Imputation) \cite{Tashiro21} extends this to multivariate time series by treating forecasting as an imputation task, utilizing a non-autoregressive approach that considers correlations across both temporal and feature dimensions. In parallel, SSSD (Structured State Space Diffusion model) \cite{alcaraz2023diffusionbasedtimeseriesimputation} explores the use of Structured State Space models within the diffusion framework to better capture long-term dependencies.

Despite these advancements, applying diffusion models to TM prediction faces specific challenges. First, most existing time-series diffusion models like CSDI and SSSD, operate directly on raw numerical sequences. While effective for general sensor data, they often struggle to capture the global spatial structure and 2D topological correlations unique to network traffic matrices. Second, more critically, current diffusion architectures typically rely on shallow conditioning mechanisms including simple concatenation of historical data and basic time embeddings. They lack a deep semantic understanding of the input history. For instance, they treat a network traffic burst caused by a recurring daily peak and a burst caused by a random anomaly as numerically similar noise patterns, without distinguishing the underlying causality. 

\subsection{LLMs}
Pretrained on massive corpora, LLMs based on the Transformer architecture have demonstrated exceptional capabilities in natural language understanding and generation. Representative models such as GPT-3 \cite{Brown20} and LLaMA \cite{touvron2023llamaopenefficientfoundation} exhibit remarkable zero-shot and few-shot reasoning abilities, allowing them to adapt to new tasks with minimal or no task-specific training. Furthermore, advancements like GPT-4 \cite{openai2024gpt4technicalreport} have pushed the boundaries of complex logical reasoning and instruction following. These models leverage vast amounts of world knowledge encoded in their parameters to infer context and patterns, motivating researchers to explore their potential beyond text processing.

In time-series analysis, for instance, researchers have creatively adapted LLMs by tokenizing numerical sequences. Zhou et al. \cite{Zhou23} treat time-series as pseudo-sentences for zero-shot forecasting, while frameworks like Time-LLM \cite{Jin24} and UniTime \cite{Liu24} have demonstrated that a frozen LLM's internal reasoning capacity can be harnessed for numerical prediction through clever input reprogramming and prompting, effectively bypassing the need for full fine-tuning.

Similarly, in computer vision, the paradigm of coupling a frozen LLM with a sensory encoder via a lightweight adapter has proven highly effective. Models like Flamingo \cite{Alayrac22} and BLIP-2 \cite{Li23} leverage this architecture for tasks like visual question answering, achieving strong few-shot performance by building a bridge between pixels and the LLM's semantic space.

However, a conspicuous gap remains when we consider the specific problem of TM prediction. While the aforementioned approaches validate the transferability of LLM capabilities to sequences and images, the unique spatio-temporal structure of a TM, viewed as a low-resolution image that simultaneously represents a dynamic graph state, places it in a category of its own. Existing time-series methods like [26, 27] primarily model univariate or multivariate sequences, lacking a native mechanism to reason about the global spatial correlations encoded in a TM. Conversely, vision-language models are optimized for natural images with rich texture and semantic content, not for the abstract, topology-dependent patterns of network traffic. This leaves the potential of LLMs as semantic reasoners for the holistic, network-wide forecasting problem fundamentally underexplored.

\textbf{Discussion:}
Overall, prior work on TM prediction has primarily focused on deterministic spatio-temporal modeling, including statistical methods and deep learning based STGNNs, as well as TM reconstruction. Recent studies have further explored diffusion-based generative models for uncertainty modeling and the use of LLMs as semantic reasoners in time-series and vision tasks. Going beyond these approaches, our proposed \textit{LEAD} formulates TM forecasting as a semantic-guided generative process that explicitly models future traffic distributions and leverages high-level reasoning from a frozen LLM for forward-looking prediction.

\begin{figure*}
	\centering
	\includegraphics[width=1\linewidth]{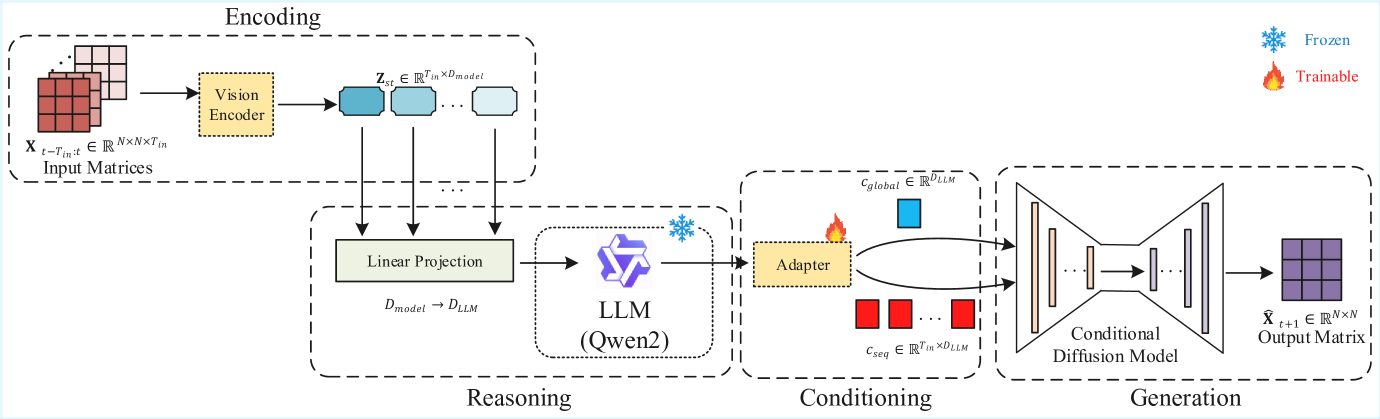}
	\caption{The Overall Structure of \textit{\textit{LEAD}}}
	\label{fig:1}
\end{figure*}

\section{Methodology}{\label{sec:methodology}}
In this section, we elaborate on the proposed framework, termed \textit{LEAD}, which synergizes the semantic reasoning capabilities of LLMs with the probability modeling power of conditional diffusion models. The core philosophy is to treat traffic prediction not merely as a numerical regression task, but as a process of semantic interpretation and generation. The overall architecture of our \textit{LEAD} is depicted in \textbf{Fig.~\ref{fig:1}}, which proceeds in four stages:

\begin{itemize}
	\item[1.] \textbf{Encoding:} Historical TMs $\textbf{X}_{1:T_{in}}$, where $T_{in}$ denotes the input length of TMs, are encoded into spatio-temporal visual tokens, preserving high-frequency details.
	\item[2.] \textbf{Reasoning:} These tokens are projected into the embedding space of a frozen LLM (Qwen2-0.5B). Through lightweight, trainable Enhanced Adapters, the model captures long-range dependencies and domain-specific traffic patterns without altering the pretrained knowledge base.
	\item[3.] \textbf{Conditioning:} The output hidden states of the LLM are disentangled into two distinct conditions: a Global Condition describing the macroscopic network state, and a Sequential Condition preserving microscopic spatio-temporal structures. 
	\item[4.] \textbf{Generation:} These conditions guide a U-Net based Diffusion Model to iteratively denoise Gaussian noise, reconstructing the future TMs $\hat{\textbf{X}}_{{T_{in}+1}:{T_{in}+T_{out}}}$, where $T_{out}$ represents the prediction horizon of TMs.
	
\end{itemize}

\subsection{Data Representation}
Network traffic matrices encapsulate sophisticated spatial relationships among nodes, presenting both an analytical challenge and an opportunity for novel representation. Drawing upon emerging paradigms that reconceptualize sequential data as visual structures \cite{Kablaoui24}, we introduce the ``\textbf{Traffic-to-Image}'' paradigm to transform these matrices into RGB images. This reformulation enables the application of advanced pretrained vision models, whose proven capacity for discerning spatial hierarchies, ranging from localized correlations to macroscopic topological arrangements, becomes directly applicable to network analytics.

The transformation is mathematically formalized through a normalized rescaling operation. For each TM $\textbf{X}_t \in \mathbb{R}^{N \times N}$ at timestep $t$, we apply min-max normalization to project element values to RGB color value $[0, 255]$:
\begin{equation}
	\textbf{X}^{G}_t(m,n) = 255×[1-\frac{\textbf{X}_t(m,n)-\text{min}(\textbf{X}_t)}{\text{max}(\textbf{X}_t)-\text{min}(\textbf{X}_t)}].
\end{equation}
In this formulation, $\text{max}(\textbf{X}_t)$ and $\text{min}(\textbf{X}_t)$ individually represent the maximum and minimum elements in the matrix $\textbf{X}_t$, and $\textbf{X}_t(m,n)$ is the element in the $m$-th row and $n$-th column of matrix $\textbf{X}_t$ $(m \leq N, n\leq N)$. With the exception of the TMs themselves, we take the first-order and second-order differences of TMs, which represent the TMs' trend and acceleration respectively, into consideration and construct the RGB image corresponding to each TM by the following formula:
\begin{equation}
	\begin{aligned}
		& \mathbf{I}_1 = [\mathbf{X}^{G}_1; \mathbf{X}^{G}_1; \mathbf{X}^{G}_1], \\
		& \mathbf{I}_2 = [\mathbf{X}^{G}_2; \mathbf{X}^{G}_2 - \mathbf{X}^{G}_1; \mathbf{X}^{G}_2 - \mathbf{X}^{G}_1], \\
		& \mathbf{I}_t = [\mathbf{X}^{G}_t; \mathbf{X}^{G}_t - \mathbf{X}^{G}_{t-1}; \mathbf{X}^{G}_t - 2\mathbf{X}^{G}_{t-1} + \mathbf{X}^{G}_{t-2}].(t \geq 3)
	\end{aligned}
\end{equation}

The transformation's principal advantage lies in its creation of a representational bridge to convolution neural networks (CNNs). By transposing network data into the visual domain, we leverage the intrinsic capabilities of vision models to identify structurally analogous patterns: intensity gradients become indicators of high-traffic links, while contiguous dark regions reveal network congestion clusters. This approach not only maintains the topological fidelity of the original data but also unlocks sophisticated feature extraction methodologies traditionally applied to visual analysis, thereby facilitating the discovery of spatially distributed patterns within network traffic flows.

\subsection{Spatio-Temporal Perception via Detail-Preserving Encoding}
Having transformed the numerical traffic data into visual representations, the subsequent challenge is to extract effective spatio-temporal features from these images.

\begin{figure}
	\centering
	\includegraphics[width=1\linewidth]{"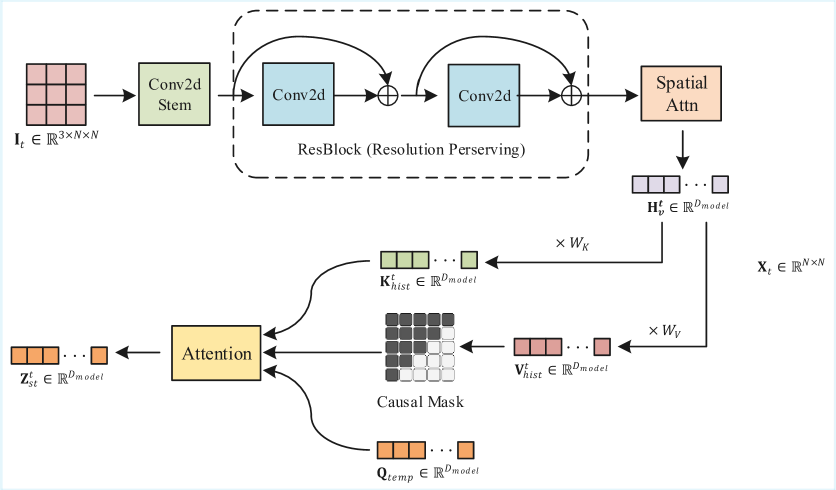"}
	\caption{Vision Encoder Structure in Detail}
	\label{fig:2}
\end{figure}

TMs differ significantly from natural images used in standard computer vision. While they possess spatial correlations (node topology), they are typically low-resolution and highly sensitive to numerical fluctuations. Standard backbones like ResNet-50 often perform excessive downsampling, obliterating critical traffic details. To mitigate this, as shown in \textbf{Fig.~\ref{fig:2}}, we design a specialized vision encoder, which prioritizes resolution preservation through a tailored stem and residual block design.

\subsubsection*{(1) Feature Extraction}
Let $\textbf{I}_t \in R^{3 \times N \times N}$ be the pixel image formed by TM at time step $t$. We employ a CNN encoder that prioritizes resolution preservation. The input first passes through a Stem layer with a small kernel size to map raw values to the feature dimension $D_v$:
\begin{equation}
	\textbf{F}_t^{stem}=\text{Conv2d}_{3 \times 3 }(\textbf{I}_t).
\end{equation}

Subsequently, the features pass through a series of Residual Blocks. Crucially, we strictly limit the stride and pooling operations to maintain the spatial resolution of the feature maps, ensuring that small-scale congestion patterns are not lost. A spatial attention mechanism is applied at the final stage to highlight critical network nodes:
\begin{equation}
	\textbf{H}_v^t=\text{SpatialAttn(ResBlock($\textbf{F}_t^{stem}$))}.
\end{equation}

\subsubsection*{(2) Causal Temporal Aggregation}
To aggregate the sequence of visual features $\textbf{H}_v^{1:T}$ into a coherent spatio-temporal representation, we introduce an enhanced temporal aggregator. Unlike bidirectional methods including Bi-LSTM (Bidirectional Long Short-Term Memory), network traffic prediction must strictly adhere to causality. We employ a Transformer-based aggregator with a causal mask, ensuring that the representation at step $t$ is based solely on the previous steps.

Specifically, the extracted visual feature sequence $\textbf{H}_v^{1:T_{in}}$ serves as the source of historical context. We project it into key and value spaces via learnable linear transformations:
\begin{equation}
	\textbf{K}_{hist}=\textbf{H}_v^{1:T_{in}}W_{K}, \textbf{V}_{hist}=\textbf{H}_v^{1:T_{in}}W_{V},
\end{equation}
where $W_K, W_V \in \mathbb{R}^{D_v \times D_{model}}$ are projection matrices. To retrieve distinct temporal patterns from this context, we define a learnable temporal query $\textbf{Q}_{temp} \in \mathbb{R}^{T \times D_{model}}$ initialized with positional embeddings. The aggregation is then computed via causal attention:
\begin{equation}
	\textbf{Z}_{st}= \text{Attention($\textbf{Q}_{temp}$, $\textbf{K}_{hist}$, $\textbf{V}_{hist} \odot \text{M}_{causal}$)},
\end{equation}
where $\textbf{M}_{causal}$ is the lower-triangular mask preventing information leakage from future time steps. This process yields the final spatio-temporal tokens $\textbf{Z}_{st} \in R^{T \times D_{model}}$, which are subsequently aligned for LLM injection.

\subsection{LLM-Based Reasoning with Adapter}
Recent studies have demonstrated that LLMs, pretrained on vast textual corpora, possess innate capabilities for sequential reasoning and pattern recognition that can transfer to time-series forecasting. We leverage a lightweight pretrained LLM, Qwen2-0.5B, as our semantic backbone. However, fine-tuning the entire LLM is computationally prohibitive and risks catastrophic forgetting. Instead, we design a ``\textbf{Frozen LLM with Trainable Adapter}'' model, which adjusts the output of the LLM via an adapter layer. 

\begin{figure}[!ht]
	\centering
	\includegraphics[width=1\linewidth]{"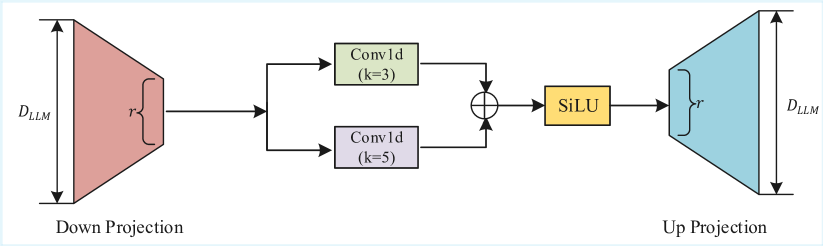"}
	\caption{Adapter Structure in Detail}
	\label{fig:3}
\end{figure}

\subsubsection*{(1) Modality Alignment}
The continuous spatio-temporal tokens $\textbf{Z}_{st}$ must be aligned with the LLM's token space. We use a linear projector $W_{proj}$ to map the visual dimension $D_{model}$ to the LLM dimension $D_{LLM}$:
\begin{equation}
	\textbf{TE}_{in}=\textbf{Z}_{st} \cdot W_{proj} +\textbf{P}_{pos},    
\end{equation}
where $\textbf{P}_{pos}$ represents learnable positional embeddings. These traffic tokens are then fed into the frozen Qwen blocks.

\subsubsection*{(2) Multi-Scale Adapter Architecture}
We insert an adapter layer parallel to the feed-forward networks (FFN) inside the Transformer blocks. As illustrated in \textbf{Fig.~\ref{fig:3}}, the adapter is designed to capture multi-scale temporal dynamics which are often absent in pure language models.

Formally, for the $l$-th layer input $\textbf{h}_{l-1}$, the output is:
\begin{equation}\mathbf{h}_{l}=\mathbf{h}_{l-1}+\mathrm{LLM}(\mathbf{h}_{l-1})+\lambda\cdot\mathrm{Adapter}(\mathbf{h}_{l-1}).\end{equation}

The adapter comprises a bottleneck structure with three key components:
\begin{itemize}
	\item \textbf{Down-Projection:} A linear layer that compresses features to a lower rank $r \ll D_{LLM}$ to minimize parameters.
	\item \textbf{Multi-Scale Convolution:} A parallel branch of 1D convolutions with kernel sizes $k \in \{3,5\}$. This allows the model to simultaneously capture short-term jitters like local traffic spikes and long-term trends like periodic patterns:
	\begin{equation}\mathbf{h}_{mid}=\mathrm{SiLU}(\mathrm{Conv}_{k=3}(\mathbf{h}_{down})+\mathrm{Conv}_{k=5}(\mathbf{h}_{down})).\end{equation}
	\item \textbf{Gated Up-Projection:} A module that restores dimensions and uses a gating mechanism to dynamically control the information flow.
\end{itemize}

This design effectively injects spatio-temporal inductive biases into the frozen LLM, transforming it into a specialized traffic reasoning engine.

\subsection{Dual-Conditioning Strategy}
Having extracted rich temporal semantic priors via the LLM, the next critical step is to map these high-level features back to the high-dimensional traffic space. We employ a diffusion model as the probabilistic decoder to reconstruct the TM, as it excels in capturing complex non-linear dependencies and mitigating the over-smoothing issues typical of deterministic regressors. However, diffusion models require robust conditioning signals to guide the denoising process effectively. Therefore, we propose a \textbf{Dual-Conditioning Strategy} that disentangles the semantic information into Global and Sequence representations.

\subsubsection*{(1) Global Condition $c_{global}$}
To encapsulate the holistic state of the network, we apply attention pooling over the LLM's output sequence $\textbf{H}_{out} \in {R^{T\times D_{LLM}}}$. Define a trainable query vector $\textbf{q}_{pool}$, the global condition $c_{global}$ is computed as a weighted sum of the sequence:
\begin{equation}
	\begin{aligned}
		& \alpha_i = \mathrm{softmax}\left(\frac{\mathbf{q}_{pool}\cdot\mathbf{H}_{out}[i]^\top}{\sqrt{d}}\right), \\[2ex]
		& c_{global} = \sum_{i=1}^T\alpha_i\mathbf{H}_{out}[i],
	\end{aligned}
\end{equation}
where $d$ denotes the dimension of the query vector.

This vector $c_{global}$ provides a style-like guidance for the diffusion model, determining the overall intensity distribution of the generated pixel images.

\subsubsection*{(2) Sequential Condition $c_{seq}$}
While the global condition captures the trend, it loses structural details. To preserve node-level interactions and temporal transitions, we project the full sequence of hidden states to the diffusion model's dimension:
\begin{equation}c_{seq}=\mathrm{LayerNorm}(\mathbf{H}_{out}\cdot W_{seq}),\end{equation}
where $\mathbf{H}_{out}$ denotes the sequence of hidden states from the LLM, and $W_{seq} \in \mathbb{R}^{D_{LLM} \times D_{model}}$ is a learnable linear projection matrix designed to align the dimension of LLM features with the latent space of the diffusion model.

This sequential condition allows the generation model to attend to specific historical moments via cross-attention, ensuring that the predicted pixel image aligns with the immediate past trajectory.  

\subsection{Conditional Diffusion Generation}
The final module is a conditional diffusion model that generates the TM $\hat{\textbf{X}}_{{T_{in}+1}:{T_{in}+T_{out}}}$. We employ a U-Net based architecture $\epsilon_{\theta}$ to approximate the noise $\epsilon$ added to the data. 

\subsubsection*{(1) Forward Diffusion Process}
Following the standard DDPM (Denoising Diffusion Probabilistic Models) formulation, we define a forward process that gradually adds Gaussian noise to the ground truth $\textbf{x}_0$ over $K$ steps. At any step $k$, the noisy data $\textbf{x}_k$ is:
\begin{equation}\mathbf{x}_k=\sqrt{\bar{\alpha}_k}\mathbf{x}_0+\sqrt{1-\bar{\alpha}_k}\epsilon,\quad\epsilon\sim\mathcal{N}(0,\mathbf{I}),\end{equation}
where $\bar{\alpha}_k$ represents the noise schedule, $x_0$ denotes the original clean data sample and $\epsilon$ is the noise injected during the forward process. This process transforms the complex traffic distribution into an isotropic Gaussian distribution.

\subsubsection*{(2) Condition Injection via U-Net}
\begin{figure*}[t]
	\centering
	\includegraphics[width=1\linewidth]{"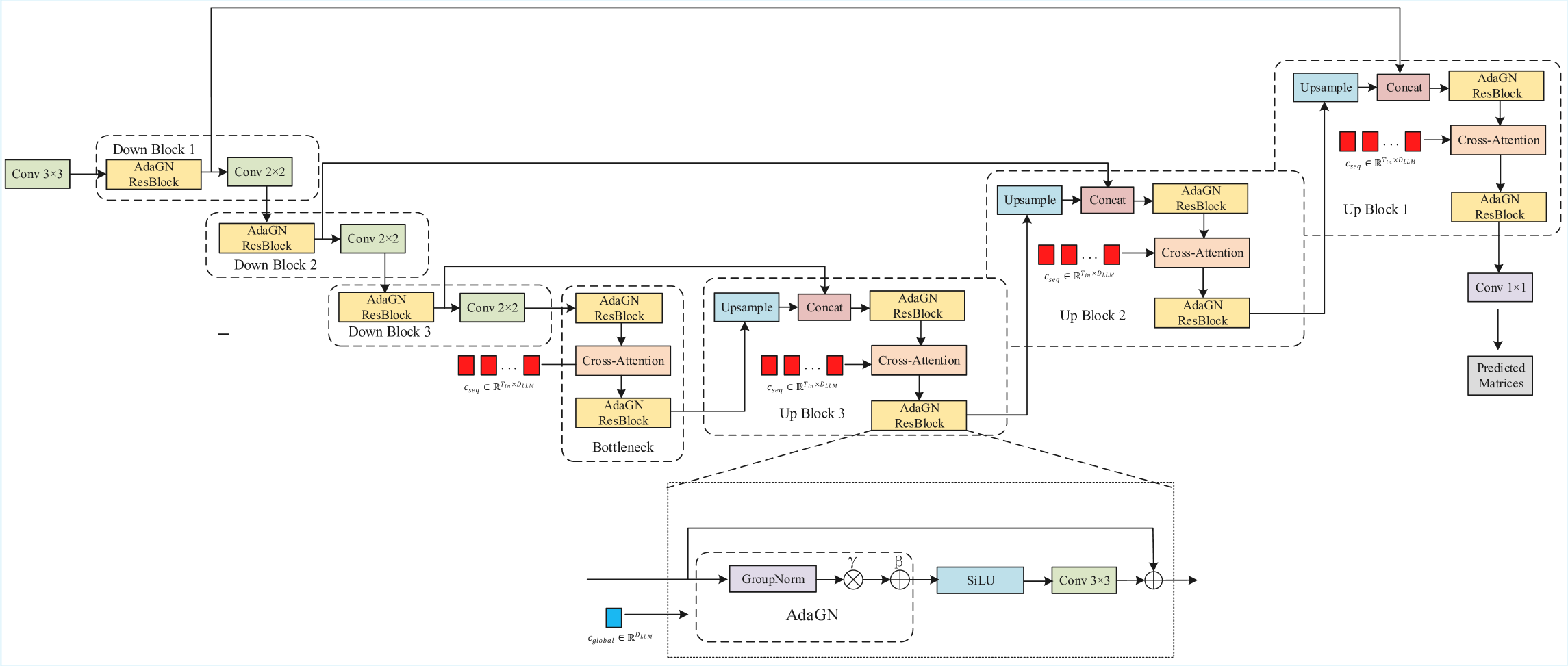"}
	\caption{Conditional U-Net Structure in Detail}
	\label{fig:4}
\end{figure*}

As depicted in \textbf{Fig.~\ref{fig:4}}, we introduce an Adaptive Group Normalization (AdaGN) mechanism to inject the global context $c_{global}$ into the U-Net backbone. AdaGN dynamically modulates the U-Net's intermediate feature maps by learning scale and shift parameters derived directly from $c_{global}$, thereby guiding the denoising trajectory with global semantic constraints: 
$\mathrm{AdaGN}(\mathbf{f},c_{global})=\gamma_{s}(c_{global})\cdot\mathrm{GroupNorm}(\mathbf{f})+\beta_{s}(c_{global})$.
Here, $\gamma_{s}$ and $\beta_{s}$ are scale and shift parameters predicted from $c_{global}$ via a shallow Multi-Layer Perceptron. This mechanism effectively shifts the distribution of the generated traffic to match the predicted global congestion level. The sequential condition $c_{seq}$ acts as the Key and Value in the cross-attention layers of the U-Net, while the intermediate feature maps of the U-Net serve as Queries.
\begin{equation}\mathrm{Attention}(Q,K,V)=\mathrm{softmax}\left(\frac{Q(\mathbf{f})\cdot K(c_{seq})^\top}{\sqrt{d}}\right)V(c_{seq}).\end{equation}
This allows the generative process to spatially align the predicted traffic peaks with the historical patterns identified by the LLM.

\subsubsection*{(3) Optimization and Inference}

The model is trained end-to-end (keeping LLM frozen, training Adapter and Diffusion U-Net) by minimizing the Mean Squared Error (MSE) between the predicted noise and the added noise:
\begin{equation}\mathcal{L}_{diff}=\mathbb{E}_{k,\mathbf{x}_0,\epsilon}\left[\|\epsilon-\epsilon_\theta(\mathbf{x}_k,k,c_{global},c_{seq})\|^2\right].\end{equation}
During inference, we employ the Denoising Diffusion Implicit Models (DDIM) sampler to accelerate the generation process, allowing us to produce the prediction $\hat{\textbf{X}}_{{T+1}:{T+T_{out}}}$ in fewer steps than the 1000-step denoising process.

\section{Experiments}{\label{sec:experiment}}
We conduct comprehensive experiments to evaluate the performance of our proposed \textit{\textit{\textit{LEAD}}} for network traffic prediction, and analyze its advantages compared to other baselines. In addition, we also conduct ablation studies, which demonstrate the effectiveness of each module in our \textit{\textit{\textit{LEAD}}}. 

\subsection{Datasets}
To rigorously evaluate our \textit{\textit{LEAD}}'s capability in handling diverse network topologies and traffic patterns, we select two real-world backbone network datasets: \textbf{Abilene} (US) and \textbf{GÉANT} (Europe). These datasets represent distinct scales and complexities, allowing us to evaluate the robustness and universality of our approach.

\textbf{The Abilene Dataset} serves as our primary benchmark for evaluating our model's capability in capturing high-frequency network traffic dynamics. Collected from the Internet2 backbone network in the United States, it comprises traffic data from 12 core nodes aggregated at 5-minute intervals. This dataset is characterized by significant traffic variability and sudden burstiness, making it ideal for testing the model's temporal sensitivity. We map the 12 nodes into a $12 \times12$ spatial grid to facilitate our diffusion process.

\textbf{The GÉANT Dataset} is introduced to assess the spatial scalability and generalization of our approach. Sourced from the pan-European research network, GÉANT connects varying National Research and Education Networks and features a significantly larger and more complex topology (23 nodes) compared to Abilene. Its traffic distribution is highly heterogeneous and spatially sparse, presenting a challenge for topology-agnostic learning. By testing on GÉANT, we validate that the ``Traffic-to-Image'' paradigm can effectively reconstruct spatial correlations across different network scales without overfitting to a specific geometric structure.

\subsection{Baselines and Metrics}
In order to comprehensively evaluate the performance of \textit{\textit{\textit{LEAD}}}, we compare it against nine representative baseline models, categorized into three groups: RNN-based models, STGNNs, and advanced multi-scale and visual learning models. 
\subsubsection*{(1) RNNs}
\
\newline
\indent We include \textbf{LSTM} and \textbf{GRU} as foundational baselines. These models focus solely on the temporal evolution of traffic sequences, treating nodes independently or as a flattened vector, thereby largely neglecting the explicit spatial topological correlations \cite{SiamiNamini19}.

\subsubsection*{(2) STGNNs}
\
\newline
\indent This category explicitly models the network topology using graph-based operations.
\begin{itemize}
	\item \textbf{DCRNN \& STGCN:} \textbf{DCRNN} integrates diffusion convolution with GRU, while \textbf{STGCN} employs graph convolutions and 1D temporal convolutions for faster inference. They both rely on predefined adjacency matrices to capture spatial dependencies \cite{li2018diffusion}, \cite{Yu18}. 
	\item \textbf{MTGNN:} Unlike the former two, MTGNN (Multivariate Time Series Graph Neural Networks) automatically learns an adaptive graph structure from data, enabling it to capture hidden relations among traffic nodes without prior topological knowledge \cite{10.1145/3394486.3403118}.
\end{itemize}

\subsubsection*{(3) Advanced Multi-Scale and Visual Learning}
\
\newline
\indent This category of approaches address complex data heterogeneity and introduce novel data representations.
\begin{itemize}
	\item \textbf{ST-SSL \& MC-STL:} \textbf{ST-SSL} employs self-supervised learning with pretext tasks to learn robust patterns resilient to noise. \textbf{MC-STL} decomposes traffic data into multiple channels to model distinct temporal dynamics separately \cite{Ji23}, \cite{Zhang23}.
	\item \textbf{M²STL:} \textbf{M²STL} (Multi-range Multi-level Spatial-Temporal Learning) is a hierarchical framework that explicitly models traffic at different temporal ranges, which capture dependencies from short-term to long-term, and spatial levels that extract features ranging from local to global scopes, effectively fusing multi-granular information \cite{Xie24}.
	\item \textbf{ViT-LSTM:} This method is particularly relevant as it shares the ``Traffic-to-Image'' paradigm with our approach. It converts TMs into image sequences and utilizes ViT (Vision Transformer) to extract spatial features, followed by LSTM for temporal modeling. By comparing with ViT-LSTM, we can directly validate the superiority of our generative diffusion backbone over traditional discriminative visual models \cite{Kablaoui24}.
\end{itemize}

The primary metrics are pixel-wise root mean squared error (RMSE) and mean absolute error (MAE), computed on the normalized color value of the first channel of RGB images. Given ground truth $X \in \mathbb{R}^{N \times N}$ and prediction $\hat{X} \in \mathbb{R}^{N \times N}$, we have:
\begin{equation}
	\begin{aligned}
		& \text{RMSE}=\sqrt[]{\frac{\sum\limits_{i=1}^{N}\sum\limits_{j=1}^N(\hat{X}(i,j)-X(i,j))^2}{N^2}},\\ &\text{MAE}=\frac{\sum\limits_{i=1}^{N}\sum\limits_{j=1}^N\mid\hat{X}(i,j)-X(i,j)\mid}{ N^2}.
	\end{aligned}
\end{equation}

\subsection{Setup and Implementation Details}
\subsubsection*{(1) Data Preprocessing and Partitioning}
\
\newline
\indent To ensure a rigorous evaluation without data leakage, we adopt a strict chronological splitting strategy for both Abilene and GÉANT datasets. The data are divided into training, validation, and testing sets with a ratio of \textbf{7:1.5:1.5}. Moreover, we employ a sliding window mechanism to generate samples. The model utilizes the historical traffic sequence of length $\bm{T_{in}=24}$ to predict the TMs at the next $\bm{T_{out}}$ time steps.

\subsubsection*{(2) Model Configuration}
\
\newline
\indent The specific hyperparameters for our $\textit{\textit{LEAD}}$ model are derived from extensive tuning:
\begin{itemize}
	\item \textbf{LLM-Adapter:} We utilize the pretrained \textbf{Qwen2-0.5B} as the semantic encoder. The LLM weights are frozen during training to preserve its generalization capability. The lightweight adapter projects LLM hidden states to a dimension of \textbf{256} and employs multi-scale convolutions with kernel sizes $\bm{k=[3,5]}$.
	\item \textbf{Conditioning:} The model integrates global context via FiLM (Feature-wise Linear Modulation) and local details via cross-attention, with a diffusion noise schedule of \textbf{1000} steps.
	\item \textbf{Diffusion Model:} The conditional U-Net backbone is configured with a base channel dimension of \textbf{32}. We employ a channel multiplier sequence of $\bm{[1,2,4]}$ with \textbf{2} Residual block groups per resolution level, ensuring sufficient depth to capture complex spatial patterns.
	
\end{itemize}

\begin{table*}[htbp]
	\centering
	
	\caption{Performance comparison of different models on the Abilene dataset, including RMSE and MAE values.}
	\label{tab:model_comparison_abilene}
	\begin{tabular}{cccccccc}
		\toprule
		\multirow{2}{*}{\textbf{Type}} & \multirow{2}{*}{\textbf{Models}} & \multicolumn{2}{c}{$\bm{T_{out}=1}$} & \multicolumn{2}{c}{$\bm{T_{out}=10}$} & \multicolumn{2}{c}{$\bm{T_{out}=20}$} \\
		\cmidrule(lr){3-4} \cmidrule(lr){5-6} \cmidrule(lr){7-8} 
		& & MAE & RMSE & MAE & RMSE & MAE & RMSE \\
		\midrule
		
		\multirow{2}{*}{RNN-based}
		& GRU & 0.0657 & 0.1313 & 0.0696 & 0.1409 & 0.0720 & 0.1417 \\
		& LSTM & 0.0654 & 0.1365 & 0.0666 & 0.1367 & 0.0735 & 0.1476 \\
		\midrule
		
		\multirow{3}{*}{ST-GNNs} 
		& DCRNN & 0.0777 & 0.1320 & 0.0832 & 0.1387 & 0.0871 & 0.1399 \\
		& STGCN & 0.1555 & 0.2323 & 0.1675 & 0.2496 & 0.1711 & 0.2533 \\
		& MTGNN & 0.0904 & 0.2003 & 0.1286 & 0.2651 & 0.1376 & 0.2781 \\
		\midrule
		
		\multirow{3}{*}{Advanced Multi-Scale} 
		& ST-SSL & 0.1174 & 0.2380 & 0.1233 & 0.2484 & 0.1244 & 0.2494 \\
		& MC-STL & 0.0702 & 0.1365 & 0.0714 & 0.1514 & 0.0737 & 0.1527 \\
		& $\text{M}^2$-STL & \underline{0.0610} & \underline{0.1198} & 0.0808 & 0.1583 & 0.0844 & 0.1669 \\
		\midrule
		
		\multirow{2}{*}{Visual Learning}
		& ViT-LSTM & 0.0707 & 0.1368 & \underline{0.0732} & \underline{0.1464} & \underline{0.0778} & \underline{0.1511} \\
		& \textbf{\textit{\textit{\textit{LEAD}}}} & \textbf{0.0491} & \textbf{0.1098} & \textbf{0.0494} & \textbf{0.1112} & \textbf{0.0504} & \textbf{0.1134} \\
		\bottomrule
	\end{tabular}
	
	\vspace{0.5cm} 
	
	\caption{Performance comparison of different models on the GÉANT dataset, including RMSE and MAE values.}
	\label{tab:model_comparison_geant}
	\begin{tabular}{cccccccc}
		\toprule
		\multirow{2}{*}{\textbf{Type}} & \multirow{2}{*}{\textbf{Models}} & \multicolumn{2}{c}{$\bm{T_{out}=1}$} & \multicolumn{2}{c}{$\bm{T_{out}=10}$} & \multicolumn{2}{c}{$\bm{T_{out}=20}$} \\
		\cmidrule(lr){3-4} \cmidrule(lr){5-6} \cmidrule(lr){7-8} 
		& & MAE & RMSE & MAE & RMSE & MAE & RMSE \\
		\midrule
		
		\multirow{2}{*}{RNN-based}
		& GRU & 0.0162 & 0.0665 & 0.0173 & 0.0659 & 0.0177 & 0.0692 \\
		& LSTM & 0.0163 & 0.0671 & 0.0166 & 0.0698 & 0.0179 & 0.0708 \\
		\midrule
		
		\multirow{3}{*}{ST-GNNs } 
		& DCRNN & 0.0235 & 0.0696 & 0.0206 & 0.0629 & 0.0163 & 0.0504 \\
		& STGCN & 0.0332 & 0.0589 & 0.0463 & 0.0903 & 0.0481 & 0.0975 \\
		& MTGNN & 0.0146 & 0.0428 & 0.0155 & 0.0715 & 0.0175 & 0.0781 \\
		\midrule
		
		\multirow{3}{*}{Advanced Multi-Scale} 
		& ST-SSL & 0.0156 & 0.0274 & 0.0168 & 0.0332 & 0.0174 & 0.0359 \\
		& MC-STL & \underline{0.0112} & 0.0310 & 0.0144 & 0.0338 & 0.0166 & 0.0390 \\
		& $\text{M}^2$-STL & \textbf{0.0088} & \underline{0.0221} & \textbf{0.0120} & \underline{0.0327} & \textbf{0.0126} & 0.0361 \\
		\midrule
		
		\multirow{2}{*}{Visual Learning}
		& ViT-LSTM & 0.0144 & 0.0351 & 0.0148 & 0.0353 & 0.0149 & \underline{0.0355} \\
		& \textbf{\textit{\textit{\textit{LEAD}}}} & 0.0119 & \textbf{0.0220} & \underline{0.0136} & \textbf{0.0228} & \underline{0.0140} & \textbf{0.0258} \\
		\bottomrule
	\end{tabular}
\end{table*}

\subsubsection*{(3) Training Protocol}
\
\newline
\indent The model is trained using the AdamW optimizer on a single NVIDIA RTX 4060 GPU. We set the initial learning rate to $3 \times 10^{-5}$ with a Cosine Annealing scheduler to stabilize convergence and the effective batch size is set to 32. The training objective is the standard MSE between the predicted noise $\epsilon_{\theta}$ and the actual Gaussian noise $\epsilon$ added to the image at timestep $t$. The model is trained for a maximum of \textbf{300} epochs. We employ early stopping with a patience of $30$ epochs based on the validation loss to prevent overfitting. 

\subsubsection*{(4) Inference}
\
\newline
\indent During the inference phase, we employ the DDIM sampling strategy with 50 steps. This acceleration significantly reduces latency compared to the full 1000-step denoising process while maintaining generation quality \cite{Song21}.

\subsection{Main Results and Analysis}

\begin{figure*}[t]
	\centering
	\subfloat[RMSE Comparison of \textit{LEAD} with STGNNs on the Abilene dataset.]{
		\includegraphics[width=0.48\linewidth]{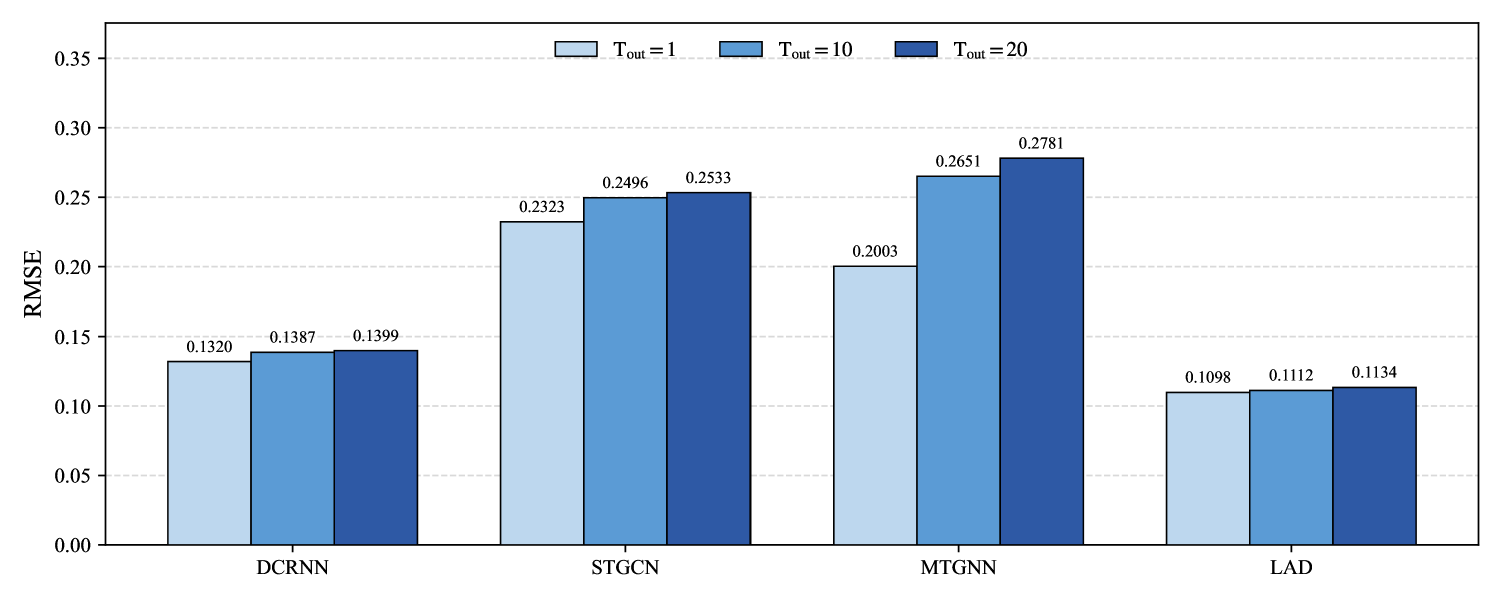}
		\label{fig:5a}
	}
	\hfill
	\subfloat[RMSE Comparison of \textit{LEAD} with STGNNs on the GÉANT dataset.]{
		\includegraphics[width=0.48\linewidth]{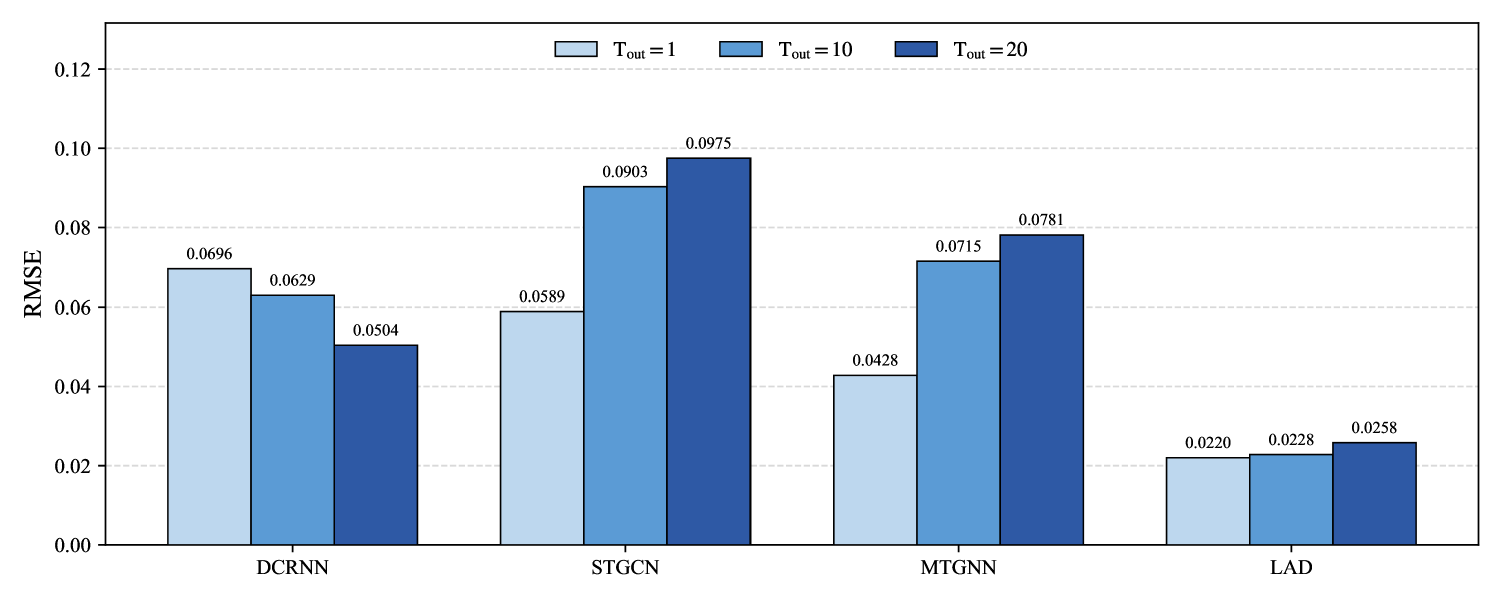}
		\label{fig:5b}
	}
	
	\vspace{0.5em}
	
	\subfloat[RMSE Comparison of \textit{LEAD} with Multi-Scale	methods on the Abilene dataset.]{
		\includegraphics[width=0.48\linewidth]{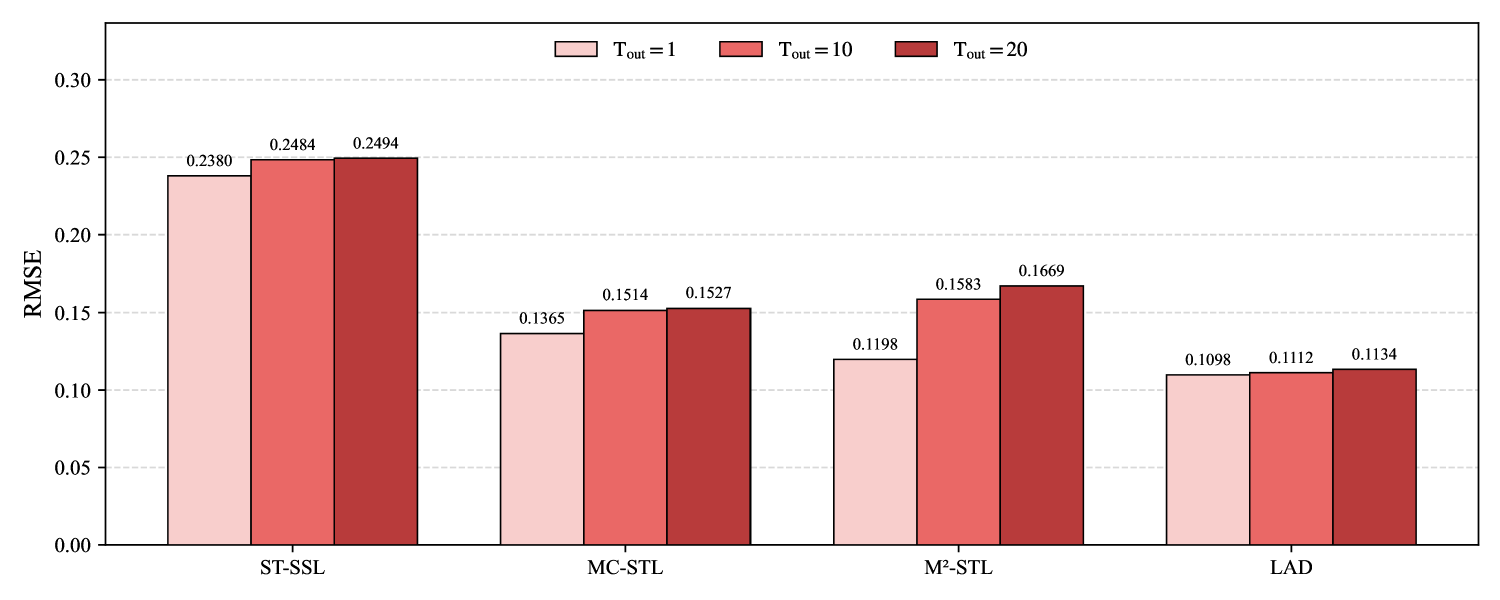}
		\label{fig:5c}
	}
	\hfill
	\subfloat[RMSE Comparison of \textit{LEAD} with Multi-Scale	methods on the GÉANT dataset.]{
		\includegraphics[width=0.48\linewidth]{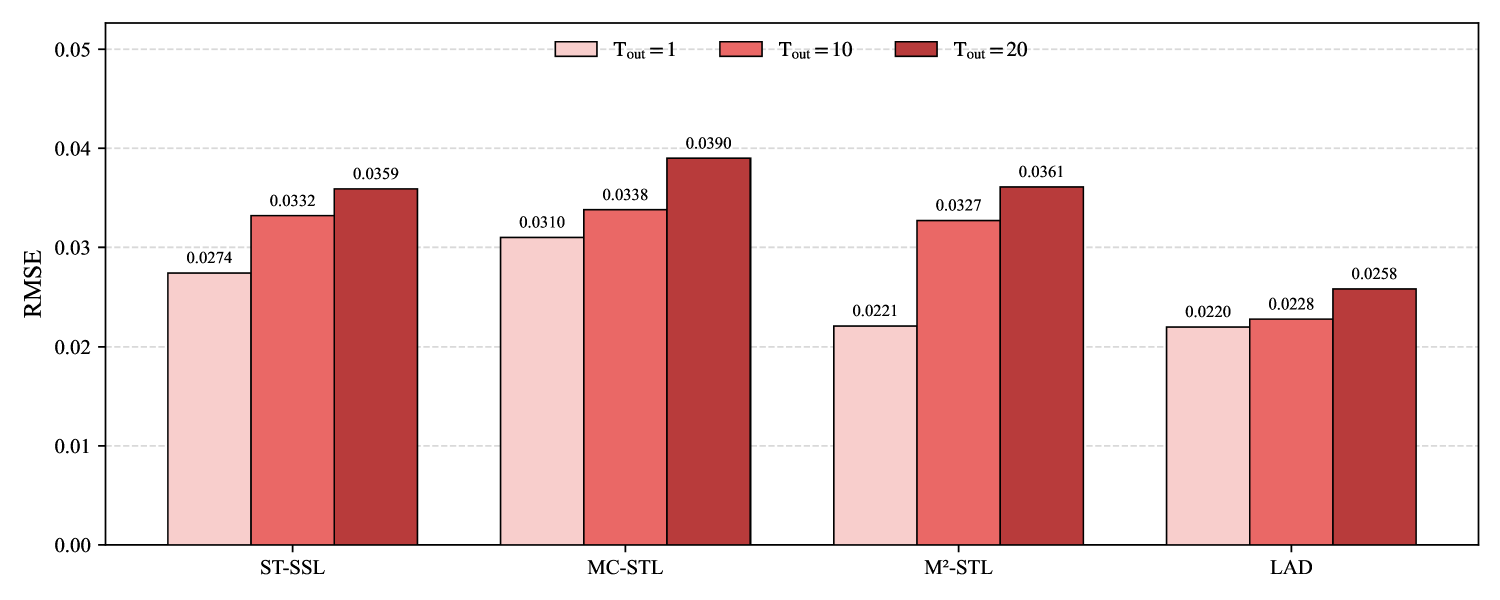}
		\label{fig:5d}
	}
	
	\caption{RMSE Comparison between \textit{\textit{LEAD}} and STGNNs \& Multi-Scale Methods}
	\label{fig:5all}
\end{figure*}

The quantitative comparison between our proposed \textit{\textit{\textit{LEAD}}} and nine baselines on the Abilene and GÉANT datasets is presented in \textbf{Table}~ \ref{tab:model_comparison_abilene} and \textbf{Table}~ \ref{tab:model_comparison_geant} respectively, covering prediction horizon from short-term ($T_{out}=1$) to long-term ($T_{out}=20$). Overall, \textit{\textit{\textit{LEAD}}} achieves outstanding performance across both datasets. To provide a more intuitive understanding of the performance divergence across different paradigms, we contrast our model with STGNNs in \textbf{Fig.~\ref{fig:5a}}-\textbf{Fig.~\ref{fig:5b}} and with multi-scale methods in \textbf{Fig.~\ref{fig:5c}}-\textbf{Fig.~\ref{fig:5d}}.

A critical observation from \textbf{Fig.~\ref{fig:5a}}-\textbf{Fig. \ref{fig:5b}} reveals distinct behaviors among GNN-based models. While DCRNN maintains relatively competitive performance among GNNs, models relying on strictly static graphs like STGCN struggle significantly with dynamic network changes, reaching an RMSE of 0.2323 on the Abilene dataset. This indicates that pre-defined adjacency matrices often fail to capture rapid traffic shifts. Even compared to MTGNN which adaptively learns graph structures, \textit{LEAD} maintains a substantial lead by achieving an RMSE of 0.1098, a 45.2\% improvement over MTGNN’s 0.2003. This superiority is further corroborated by the long-term stability analysis in Table~\ref{tab:model_comparison_abilene}. While MTGNN suffers from severe error accumulation where its MAE deteriorates by 52.2\% from the first to the twentieth time step, \textit{LEAD} demonstrates exceptional robustness with a negligible error growth of only 2.6\%. This confirms that our ``Traffic-to-Image'' paradigm effectively decouples long-term prediction accuracy from short-term history, avoiding the cascading errors inherent in recursive graph convolutions.

By comparing \textit{\textit{LEAD}} with advanced decomposition-based models like M²STL and MC-STL, we can validate the effectiveness of the LLM-Adapter module. Models such as MC-STL typically rely on explicit mathematical feature engineering, such as decomposing traffic into trend, seasonality, and residue components. While effective for regular patterns, these inductive biases may lack the flexibility to interpret non-stationary anomalies that deviate from standard periodic cycles. Instead of manual decomposition, \textit{\textit{LEAD}} leverages the pretrained knowledge of Qwen2-0.5B as a semantic prior. The LLM possesses inherent capabilities in sequence reasoning and causal inference \cite{Chang25}, enabling \textit{\textit{\textit{LEAD}}} to be superior. As shown in \textbf{Fig.~\ref{fig:5c}}, this allows \textit{\textit{LEAD}} to surpass the performance of M²STL which yields an RMSE of 0.1198 on Abilene. The advantage is even more pronounced in the sparse environment of the GÉANT dataset shown in \textbf{Fig.~\ref{fig:5d}}, where \textit{LEAD} achieves the lowest RMSE of 0.0220. Moreover, this superiority is sustained across extended prediction horizons. While fixed decomposition rules often fail to adapt to pattern drifts over long intervals, the semantic priors of \textit{\textit{LEAD}} ensure consistent accuracy from short-term to long-term horizons, avoiding the performance degradation commonly observed in static feature engineering approaches.

The comparison between \textit{\textit{LEAD}} and ViT-LSTM, the leading baseline in the visual learning category, offers critical insights into the efficacy of our approach. While both models leverage visual representations of traffic matrices, their performance divergence highlights the distinct mechanisms of the generative diffusion paradigm versus the discriminative regression framework. On the Abilene dataset which is characterized by sharp traffic spikes, \textit{\textit{LEAD}} demonstrates a substantial improvement by reducing the MAE by 43.99\% compared to ViT-LSTM when $T_{out}=1$  as shown in \textbf{Table}~\ref{tab:model_comparison_abilene}. Regression models like ViT-LSTM typically minimize pixel-wise error, which mathematically drives the prediction towards the conditional mean of future states. In bursty scenarios, this often results in an over-smoothing effect, where the model predicts average values to minimize penalty, thereby attenuating extreme peaks \cite{Ji23}. On the other hand, as a probabilistic generative model, \textit{\textit{LEAD}} approximates the full conditional distribution $p_{\theta}(x_{t+1}| x_{0:t})$. Instead of averaging potential futures, it samples high-fidelity realizations from this distribution. This capability allows \textit{\textit{LEAD}} to reconstruct sharp, high-frequency traffic structures that are often smoothed out by regression models. On the GÉANT dataset, this advantage is further quantified by the significant gap in RMSE. While ViT-LSTM performs adequately on general trends, its higher RMSE of 0.0351 compared to the 0.0220 achieved by \textit{\textit{LEAD}} reveals a limitation in capturing outliers. Since RMSE penalizes large errors quadratically, the superior performance of \textit{\textit{LEAD}} on this metric indicates higher precision in predicting peak congestion and anomalies.

\subsection{Ablation Studies}
To verify the effectiveness of the key components within the \textit{\textit{\textit{LEAD}}} model, we conduct ablation studies on the Abilene and GÉANT datasets as summarized in \textbf{Table}~\ref{tab:ablation}. We evaluate five variants by selectively removing specific modules:\newline
\indent \textbf{1. w/o Multi-Scale Convolution (MSC):} Replace the parallel branch of 1D convolutions with a simple linear projection.

\indent \textbf{2. w/o Global Condition $c_{global}$:} Remove the global context conditioning.

\indent \textbf{3. w/o Sequential Condition $c_{seq}$:} Remove the historical sequence context conditioning.

\indent \textbf{4. w/o LLM:} Remove the Qwen2-0.5B semantic encoder entirely with a simple linear layer.

\indent \textbf{5. Grayscale Graph Representation:} Barely use the grayscale image, rather than the RGB images, corresponding to $\mathbf{X}^{G}_t$ to represent all the matrices.

The performance hierarchy clearly demonstrates that while all components contribute to the final accuracy, the LLM semantic prior plays the most decisive role. The most striking observation from Table~\ref{tab:ablation} is the catastrophic performance degradation upon removing the LLM component. On the Abilene dataset, the variant without the LLM exhibits an RMSE of 0.2142. Similarly, on the GÉANT dataset, the error escalates significantly to an RMSE of 0.0538 compared to the optimal 0.0220 achieved by the full model. This consistent gap fundamentally validates the core hypothesis of our work. Standard diffusion models which rely solely on statistical image patterns often fail to grasp the complex and non-stationary dynamics of network traffic. The LLM acts as a semantic reasoner that injects crucial world knowledge and causal reasoning into the generation process. Without this high-level guidance, the model struggles to distinguish meaningful traffic bursts from random noise, causing a collapse in predictive fidelity regardless of whether the traffic is dense or sparse.

Beyond the semantic core, contextual conditioning mechanisms function as essential boundary constraints that underpin generation stability. Empirical results demonstrate that ablating the global condition $c_{global}$ leads to RMSEs of 0.1164 on Abilene and 0.0228 on GÉANT, which are consistently higher than those incurred by omitting the sequential condition $c_{seq}$, thereby underscoring the critical role of preserving a macroscopic network perspective for spatial consistency. Concurrently, the sequential condition grounds the generative trajectory in historical dynamics. Moreover, eliminating the multi-scale convolution within the adapter incurs a measurable performance decline, with RMSE increasing to 0.1130 on Abilene and 0.0242 on GÉANT. Although this degradation is less pronounced than that caused by removing the semantic core, it corroborates the necessity of hierarchical feature processing. Given that network traffic exhibits intrinsic variability across scales from micro-bursts to macro-trends, the adapter effectively aligns high-level semantic representations with corresponding resolution levels in the latent space, thereby enhancing the fidelity of fine-grained structural reconstruction.

Finally, we investigate the impact of our visual encoding strategy by replacing the multi-channel RGB representation with a single-channel grayscale input. As detailed in Table~\ref{tab:ablation}, this simplification leads to a marked deterioration in prediction accuracy. The RMSE on the Abilene dataset rises significantly to 0.1658, and a similar trend is observed on the GÉANT dataset where the RMSE increases to 0.0384. This performance drop highlights that the proposed three-channel configuration, which explicitly encodes traffic intensity alongside its first-order and second-order derivatives, provides critical dynamic information that a simple intensity map lacks. By flattening these distinct temporal properties into a single grayscale layer, the model is deprived of the explicit velocity and acceleration cues necessary for distinguishing complex flow evolution. Consequently, the diffusion backbone loses the multi-dimensional feature space required to reconstruct sharp traffic bursts, proving that the RGB mapping is not merely a formatting choice but a vital feature engineering step for capturing high-frequency network traffic dynamics.

\begin{table}[htbp]
	\centering
	\caption{Ablation studies of different components, showing RMSE and MAE values on the Abilene and GÉANT datasets.}
	\label{tab:ablation}
	\renewcommand{\arraystretch}{1.2} 
	\setlength{\tabcolsep}{12pt}      
	
	\begin{tabular}{c|c|cc}
		\toprule[1.5pt] 
		Dataset & Method & MAE & RMSE \\
		\midrule[1pt]   
		
		\multirow{6}{*}{Abilene} 
		& w/o MSC & \underline{0.0522} & \underline{0.1119} \\
		& w/o $c_{global}$  & 0.0567 & 0.1164 \\
		& w/o $c_{seq}$  & 0.0523 & 0.1141 \\
		& w/o LLM & 0.1219 & 0.2142 \\
		& Grayscale Graph & 0.0817 & 0.1658 \\
		& \textbf{Ours} & \textbf{0.0494} & \textbf{0.1098} \\
		
		\midrule 
		\multirow{6}{*}{G\'EANT} 
		& w/o MSC & 0.0166 & \underline{0.0232} \\
		& w/o $c_{global}$  & \underline{0.0147} & 0.0269 \\
		& w/o $c_{seq}$  & 0.0173 & 0.0285 \\
		& w/o LLM & 0.0497 & 0.0716 \\
		& Grayscale Graph & 0.0187 & 0.0353 \\
		& \textbf{Ours} & \textbf{0.0120} & \textbf{0.0220} \\
		
		\bottomrule[1.5pt] 
	\end{tabular}
\end{table}

\section{Conclusion}{\label{sec:conclusion}}
In this paper, we propose \textit{\textit{LEAD}}, a competitive model that synergizes the semantic reasoning capability of a frozen LLM (via parameter-efficient adapters) with the probabilistic modeling power of conditional diffusion for TM forecasting. By adopting a generative perspective rather than purely regressive prediction, \textit{LEAD} mitigates the over-smoothing issue of conventional baselines and reduces dependence on static graph topologies. Experiments on Abilene and GÉANT validate that semantic-guided diffusion achieves higher-fidelity forecasts under bursty and non-stationary dynamics. These results support foundation-model-empowered predictive networking as a practical building block for AI-native edge operation, especially when efficient adaptation and deployment under resource constraints are required.

Despite these promising results, two limitations remain to be addressed. First, the iterative denoising mechanism inherently limits inference efficiency compared to one-step baselines. Second, the reliance on a frozen, lightweight LLM, selected to maintain computational balance, constrains deep domain adaptation, leaving the full potential of LLM-driven reasoning in network physics largely underexplored.

Therefore, future work will focus on two main directions:\newline 
\indent (1) improving inference speed through sampling acceleration techniques such as consistency distillation; \newline
\indent (2) unlocking the full potential of the semantic core by exploring parameter-efficient fine-tuning like Low-Rank Adaptation (LoRA) \cite{Hu22} on larger-scale models to better capture intricate network dynamics. We also aim to extend this framework to broader tasks, such as network anomaly detection and root cause analysis.

\section*{Acknowledgment}
This paper is funded by Beijing University of Posts and Telecommunications-China Mobile Communications Group Co. Ltd. Joint Institute (Grant No. C2025001-A2025273).


\bibliographystyle{IEEEtran}
\bibliography{ref} 

\begin{thebibliography}{10}
\providecommand{\url}[1]{#1}
\csname url@samestyle\endcsname
\providecommand{\newblock}{\relax}
\providecommand{\bibinfo}[2]{#2}
\providecommand{\BIBentrySTDinterwordspacing}{\spaceskip=0pt\relax}
\providecommand{\BIBentryALTinterwordstretchfactor}{4}
\providecommand{\BIBentryALTinterwordspacing}{\spaceskip=\fontdimen2\font plus
\BIBentryALTinterwordstretchfactor\fontdimen3\font minus
  \fontdimen4\font\relax}
\providecommand{\BIBforeignlanguage}[2]{{%
\expandafter\ifx\csname l@#1\endcsname\relax
\typeout{** WARNING: IEEEtran.bst: No hyphenation pattern has been}%
\typeout{** loaded for the language `#1'. Using the pattern for}%
\typeout{** the default language instead.}%
\else
\language=\csname l@#1\endcsname
\fi
#2}}
\providecommand{\BIBdecl}{\relax}
\BIBdecl

\bibitem{Liu25_Survey_GenAI_6G}
Y.-J. Liu, X.~Xu, H.~Du, R.~Zhang, B.~Cao, D.~I. Kim, A.~Jamalipour, K.~B.
  Letaief, R.~Tafazolli, and G.~Feng, ``A survey of integrating generative
  artificial intelligence and {6G} mobile services: Architectures, solutions,
  technologies, and outlooks,'' \emph{{IEEE} Transactions on Cognitive
  Communications and Networking}, vol.~11, no.~3, pp. 1355--1378, 2025.

\bibitem{Qin25_GenAI_IntentDriven}
X.~Qin, M.~Sun, J.~Dai, P.~Ma, Y.~Cao, J.~Zhang, J.~Wang, D.~Niyato, P.~Zhang,
  and X.~Xu, ``Generative {AI} meets wireless networking: An interactive
  paradigm for intent-driven communications,'' \emph{{IEEE} Transactions on
  Cognitive Communications and Networking}, vol.~11, no.~3, pp. 1349--1367,
  2025.

\bibitem{Zhang25_GC_FL_Traffic}
H.~Zhang, S.~Zhou, and Z.~Niu, ``Gradient compression and correlation driven
  federated learning for wireless traffic prediction,'' \emph{{IEEE}
  Transactions on Cognitive Communications and Networking}, vol.~11, no.~4, pp.
  2247--2260, 2025.

\bibitem{Vardi96}
Y.~Vardi, ``Network tomography: Estimating source-destination traffic
  intensities from link data,'' \emph{Journal of the American Statistical
  Association}, vol.~91, no. 433, pp. 365--377, 1996.

\bibitem{Sun25_TME_LinkLoadSampling}
W.~Sun, Q.~Chen, X.~Song, E.~Bertino, and C.~Wang, ``Network traffic matrix
  estimation based on link loads sampling,'' \emph{{IEEE} Transactions on
  Network Science and Engineering}, vol.~12, no.~3, pp. 1524--1538, 2025.

\bibitem{Lv24_KET_FL}
Z.~Lv, Y.~Liu, X.~Wang, P.~Gao, Z.~Li, and Y.~Jiang, ``A knowledge-enhanced
  transformer-fl method for fault root cause localization,'' in
  \emph{Proceedings of the 33rd ACM International Conference on Information and
  Knowledge Management (CIKM)}, 2024, pp. 1607--1616.

\bibitem{Zhang24_AIKF_TMMonitoring}
Q.~Zhang and S.~Pan, ``An {AI}-augmented {K}alman filter approach to monitoring
  network traffic matrix,'' \emph{{IEEE} Transactions on Network Science and
  Engineering}, vol.~11, no.~3, pp. 2426--2437, 2024.

\bibitem{Liu25_SAMS_GNN}
Y.~Liu, Z.~Feng, Z.~Han, and P.~Zhang, ``{SAMS-GNN}: Self-adaptive multi-scale
  graph neural network for multi-band spectrum prediction,'' \emph{{IEEE}
  Transactions on Cognitive Communications and Networking}, vol.~11, no.~3, pp.
  1445--1458, 2025.

\bibitem{Kablaoui24}
R.~Kablaoui, I.~Ahmad, S.~Abed, and M.~Awad, ``Network traffic prediction by
  learning time series as images,'' \emph{Engineering Science and Technology,
  an International Journal}, vol.~55, p. 101754, 2024.

\bibitem{He16}
K.~He, X.~Zhang, S.~Ren, and J.~Sun, ``Deep residual learning for image
  recognition,'' in \emph{Proceedings of the IEEE Conference on Computer Vision
  and Pattern Recognition (CVPR)}, 2016, pp. 770--778.

\bibitem{Ho20}
J.~Ho, A.~Jain, and P.~Abbeel, ``Denoising diffusion probabilistic models,'' in
  \emph{Advances in Neural Information Processing Systems (NeurIPS)}, vol.~33,
  2020, pp. 6840--6851.

\bibitem{10835411}
X.~Yuan, Y.~Qiao, Z.~Wei, Z.~Zhang, M.~Li, P.~Zhao, R.~Hu, and W.~Li,
  ``Diffusion models meet network management: Improving traffic matrix analysis
  with diffusion-based approach,'' \emph{IEEE Transactions on Network and
  Service Management}, vol.~22, no.~2, pp. 1259--1275, 2025.

\bibitem{Zhang24}
R.~Zhang, J.~Han, C.~Liu, A.~Zhou, P.~Lu, Y.~Qiao, H.~Li, and P.~Gao,
  ``{LLaMA-Adapter}: Efficient fine-tuning of large language models with
  zero-init attention,'' in \emph{International Conference on Learning
  Representations (ICLR)}, 2024.

\bibitem{Groschwitz94}
N.~K. Groschwitz and G.~C. Polyzos, ``A time series model of long-term {NSFNET}
  backbone traffic,'' in \emph{Proceedings of IEEE INFOCOM}, March 1994, pp.
  1400--1407.

\bibitem{Duker24}
M.-C. D{\"u}ker, D.~S. Matteson, R.~S. Tsay, and I.~Wilms, ``Vector
  autoregressive moving average models: A review,'' \emph{arXiv preprint
  arXiv:2406.19702}, 2024.

\bibitem{11185220}
Y.~Liu, M.~Feng, L.~Xiao, Z.~Li, T.~Bi, and T.~Jiang, ``Lstm-based packet loss
  prediction for adaptive polar coding in quic enabled wireless networks,''
  \emph{IEEE Transactions on Cognitive Communications and Networking}, vol.~12,
  pp. 2844--2857, 2026.

\bibitem{li2018diffusion}
\BIBentryALTinterwordspacing
Y.~Li, R.~Yu, C.~Shahabi, and Y.~Liu, ``Diffusion convolutional recurrent
  neural network: Data-driven traffic forecasting,'' in \emph{International
  Conference on Learning Representations}, 2018. [Online]. Available:
  \url{https://openreview.net/forum?id=SJiHXGWAZ}
\BIBentrySTDinterwordspacing

\bibitem{Yu18}
B.~Yu, H.~Yin, and Z.~Zhu, ``Spatio-temporal graph convolutional networks: A
  deep learning framework for traffic forecasting,'' in \emph{Proceedings of
  the 27th International Joint Conference on Artificial Intelligence (IJCAI)},
  2018, pp. 3634--3640.

\bibitem{VanDenOord16}
A.~van~den Oord, S.~Dieleman, H.~Zen, K.~Simonyan, O.~Vinyals, A.~Graves,
  N.~Kalchbrenner, A.~Senior, and K.~Kavukcuoglu, ``{WaveNet}: A generative
  model for raw audio,'' in \emph{9th ISCA Speech Synthesis Workshop (SSW)},
  2016, p. 125.

\bibitem{Bai20}
L.~Bai, L.~Yao, C.~Li, X.~Wang, and C.~Wang, ``Adaptive graph convolutional
  recurrent network for traffic forecasting,'' in \emph{Advances in Neural
  Information Processing Systems (NeurIPS)}, vol.~33, 2020, pp.
  17\,804--17\,815.

\bibitem{Cui22}
Y.~Cui, K.~Zheng, D.~Cui, J.~Xie, L.~Deng, F.~Huang, and X.~Zhou, ``{METRO}: A
  generic graph neural network framework for multivariate time series
  forecasting,'' \emph{Proceedings of the VLDB Endowment}, vol.~15, no.~2, pp.
  224--236, 2022.

\bibitem{Zhao23}
K.~Zhao, C.~Guo, Y.~Cheng, P.~Han, M.~Zhang, and B.~Yang, ``Multiple time
  series forecasting with dynamic graph modeling,'' \emph{Proceedings of the
  VLDB Endowment}, vol.~17, no.~4, pp. 753--765, 2023.

\bibitem{Ji23}
J.~Ji, J.~Wang, C.~Huang, J.~Wu, B.~Xu, Z.~Wu, J.~Zhang, and Y.~Zheng,
  ``Spatio-temporal self-supervised learning for traffic flow prediction,'' in
  \emph{Proceedings of the AAAI Conference on Artificial Intelligence (AAAI)},
  vol.~37, no.~4, 2023, pp. 4377--4385.

\bibitem{Zhang23}
X.~Zhang, Y.~Gong, X.~Dong, X.~Wu, and X.~Qin, ``Mask- and contrast-enhanced
  spatio-temporal learning for urban flow prediction,'' in \emph{Proceedings of
  the 32nd ACM International Conference on Information and Knowledge Management
  (CIKM)}, 2023, pp. 3300--3309.

\bibitem{11114796}
C.~Li, L.~Feng, W.~Li, and F.~Zhou, ``Long-term traffic flow prediction: A
  knowledge-driven graph attention spatio-temporal network,'' \emph{IEEE
  Transactions on Network and Service Management}, vol.~22, no.~5, pp.
  4206--4221, 2025.

\bibitem{Rasul21}
K.~Rasul, C.~Seward, I.~Schuster, and R.~Vollgraf, ``{TimeGrad}: Autoregressive
  denoising diffusion models for multivariate probabilistic time series
  forecasting,'' in \emph{Proceedings of the 38th International Conference on
  Machine Learning (ICML)}, 2021, pp. 8857--8868.

\bibitem{Tashiro21}
Y.~Tashiro, J.~Song, Y.~Song, and S.~Ermon, ``{CSDI}: Conditional score-based
  diffusion models for probabilistic time series imputation,'' in
  \emph{Advances in Neural Information Processing Systems (NeurIPS)}, vol.~34,
  2021, pp. 24\,804--24\,816.

\bibitem{alcaraz2023diffusionbasedtimeseriesimputation}
\BIBentryALTinterwordspacing
J.~M.~L. Alcaraz and N.~Strodthoff, ``Diffusion-based time series imputation
  and forecasting with structured state space models,'' 2023. [Online].
  Available: \url{https://arxiv.org/abs/2208.09399}
\BIBentrySTDinterwordspacing

\bibitem{Brown20}
T.~B. Brown, B.~Mann, N.~Ryder, M.~Subbiah, J.~Kaplan, P.~Dhariwal
  \emph{et~al.}, ``Language models are few-shot learners,'' in \emph{Advances
  in Neural Information Processing Systems (NeurIPS)}, vol.~33, 2020, pp.
  1877--1901.

\bibitem{touvron2023llamaopenefficientfoundation}
\BIBentryALTinterwordspacing
H.~Touvron, T.~Lavril, G.~Izacard, X.~Martinet, M.-A. Lachaux, T.~Lacroix,
  B.~Rozière, N.~Goyal, E.~Hambro, F.~Azhar, A.~Rodriguez, A.~Joulin,
  E.~Grave, and G.~Lample, ``Llama: Open and efficient foundation language
  models,'' 2023. [Online]. Available: \url{https://arxiv.org/abs/2302.13971}
\BIBentrySTDinterwordspacing

\bibitem{openai2024gpt4technicalreport}
\BIBentryALTinterwordspacing
O.~et.al., ``Gpt-4 technical report,'' 2024. [Online]. Available:
  \url{https://arxiv.org/abs/2303.08774}
\BIBentrySTDinterwordspacing

\bibitem{Zhou23}
T.~Zhou, P.~Niu, X.~Wang, L.~Sun, and R.~Jin, ``One fits all: Power general
  time series analysis by pretrained {LM},'' in \emph{Advances in Neural
  Information Processing Systems (NeurIPS)}, vol.~36, 2023.

\bibitem{Jin24}
M.~Jin, S.~Wang, L.~Ma, Z.~Chu, J.~Y. Zhang, X.~Shi, P.-Y. Chen, Y.~Liang,
  Y.-F. Li, S.~Pan, and Q.~Wen, ``{Time-LLM}: Time series forecasting by
  reprogramming large language models,'' in \emph{International Conference on
  Learning Representations (ICLR)}, 2024.

\bibitem{Liu24}
X.~Liu, J.~Hu, Y.~Li, S.~Diao, Y.~Liang, B.~Hooi, and R.~Zimmermann,
  ``{UniTime}: A language-empowered unified model for cross-domain time series
  forecasting,'' in \emph{Proceedings of the ACM Web Conference (WWW)}, 2024,
  pp. 4321--4332.

\bibitem{Alayrac22}
J.-B. Alayrac, J.~Donahue, P.~Luc, A.~Miech, I.~Barr, Y.~Hasson, K.~Lenc,
  A.~Mensch, K.~Millican, M.~Reynolds \emph{et~al.}, ``Flamingo: a visual
  language model for few-shot learning,'' in \emph{Advances in Neural
  Information Processing Systems (NeurIPS)}, vol.~35, 2022, pp.
  23\,716--23\,736.

\bibitem{Li23}
J.~Li, D.~Li, S.~Savarese, and S.~Hoi, ``{BLIP-2}: Bootstrapping language-image
  pre-training with frozen image encoders and large language models,'' in
  \emph{Proceedings of the 40th International Conference on Machine Learning
  (ICML)}, 2023, pp. 19\,730--19\,742.

\bibitem{SiamiNamini19}
S.~Siami-Namini, N.~Tavakoli, and A.~S. Namin, ``The performance of {LSTM} and
  {BiLSTM} in forecasting time series,'' in \emph{Proceedings of the IEEE
  International Conference on Big Data (Big Data)}, 2019, pp. 3285--3292.

\bibitem{10.1145/3394486.3403118}
\BIBentryALTinterwordspacing
Z.~Wu, S.~Pan, G.~Long, J.~Jiang, X.~Chang, and C.~Zhang, ``Connecting the
  dots: Multivariate time series forecasting with graph neural networks,'' ser.
  KDD '20.\hskip 1em plus 0.5em minus 0.4em\relax New York, NY, USA:
  Association for Computing Machinery, 2020, p. 753–763. [Online]. Available:
  \url{https://doi.org/10.1145/3394486.3403118}
\BIBentrySTDinterwordspacing

\bibitem{Xie24}
R.~Xie, J.~Wen, X.~Chen, K.~Xie, W.~Liang, N.~Xiong, D.~Zhang, G.~Xie, and
  K.~Li, ``{M2STL}: Multi-range multi-level spatial-temporal learning model for
  network traffic prediction,'' \emph{IEEE Transactions on Network Science and
  Engineering}, vol.~11, no.~5, pp. 4315--4329, 2024.

\bibitem{Song21}
J.~Song, C.~Meng, and S.~Ermon, ``Denoising diffusion implicit models,'' in
  \emph{International Conference on Learning Representations (ICLR)}, 2021.

\bibitem{Chang25}
C.~Chang, W.-Y. Wang, W.-C. Peng, and T.-F. Chen, ``{LLM4TS}: Aligning
  pre-trained {LLMs} as data-efficient time-series forecasters,'' \emph{ACM
  Transactions on Intelligent Systems and Technology}, 2025.

\bibitem{Hu22}
E.~J. Hu, Y.~Shen, P.~Wallis, Z.~Allen-Zhu, Y.~Li, S.~Wang, L.~Wang, and
  W.~Chen, ``{LoRA}: Low-rank adaptation of large language models,'' in
  \emph{International Conference on Learning Representations (ICLR)}, 2022.

\end{thebibliography}

\end{document}